\documentclass{article} 
\usepackage{iclr2027_conference,times}

\usepackage{amsmath,amsfonts,bm}

\def\eqref#1{equation~\ref{#1}}

\def\1{\bm{1}}

\DeclareMathAlphabet{\mathsfit}{\encodingdefault}{\sfdefault}{m}{sl}
\SetMathAlphabet{\mathsfit}{bold}{\encodingdefault}{\sfdefault}{bx}{n}

\usepackage{graphicx}
\usepackage{hyperref}
\usepackage{url}
\usepackage{subcaption} 
\usepackage{wrapfig} 
\usepackage{booktabs} 
\usepackage{algorithm}
\usepackage{algpseudocode}
\usepackage{amssymb}
\usepackage{placeins}
\usepackage{xspace}
\usepackage[utf8]{inputenc}


\newcommand{\NAME}{ReLaG\xspace}

\title{\NAME: A Scalable Framework Generalizing Random Splits to Data with Latent Relations}

\author{Anthony Lavertu \& Jacob Côté \\
Department of Computer Science\\
Université Laval\\
Québec, QC, Canada \\
\texttt{\{anthony.lavertu.1,jacob.cote.3\}@ulaval.ca} \\
\And
Sophie Gobeil \\
Department of Biochemistry, Microbiology \\
and Bioinformatics \\
Université Laval \\
Québec, QC, Canada \\
\texttt{sophie.gobeil@bcm.ulaval.ca} \\
\And
Jacques Corbeil \\
Department of Molecular Medicine \\
Université Laval \\
Québec, QC, Canada \\
\texttt{jacques.corbeil@fmed.ulaval.ca} \\
\And
Isabeau Prémont-Schwarz \& Pascal Germain\\
Department of Computer Science \\
Université Laval \\
Québec, QC, Canada \\
\texttt{isabeau.premont-schwarz@ift.ulaval.ca} \\
\texttt{pascal.germain@ift.ulaval.ca}
}
\newcommand{\ourparagraph}[1]{\textbf{#1\ }}

\iclrfinalcopy 
\begin{document}

\maketitle
\lhead{Under review as a conference paper at ICLR 2027}

\begin{abstract}
Random splitting can yield non-independent train--test subsets when a dataset contains related samples, as is common in certain applications such as biochemical studies. This leads to overly optimistic generalization estimates. Here, we introduce \NAME, a modality-agnostic framework that models sample relatedness through a hierarchical latent-variable process and infers groups of related samples using proximity graphs and community detection to produce independent train--test subsets. Across molecular and protein datasets, \NAME matches existing relation-aware methods while scaling substantially better, enabling splits at previously impractical dataset sizes. We further introduce a label-free procedure that adapts the splitting resolution to production data, aligning evaluation with the intended deployment setting. \NAME's inferred groups provide a cheap estimate of effective dataset size, enabling diversity-aware dataset scaling. \NAME is open source and can be installed with \texttt{pip install relag}.
\end{abstract}


\section{Introduction}
Standard machine-learning evaluation commonly relies on random train--test splits, which assume that samples are independently and identically distributed (i.i.d.)~\citep{og_random_split}. This assumption is frequently violated in real-world datasets, where samples occur in related groups, as observed in domains ranging from epidemiology and econometrics to the social sciences~\citep{paper-clustered_data}. The issue is particularly pronounced in biochemical machine learning: evolutionary processes, combinatorial screening, and dataset bias lead to datasets that contain complex dependencies between samples~\citep{paper-guiding_questions_leakage}.

When a random split places related samples in both the training and test sets, a model can achieve high test performance by memorizing training examples rather than by learning to generalize to genuinely independent samples. This train--test leakage can therefore lead to overly optimistic generalization estimates~\citep{paper-rs_generic_leakage}. Such leakage has been documented across biochemical prediction tasks, including drug--target interaction prediction~\citep{paper-dti_random_split_memorization}, reaction prediction~\citep{paper-reaction_prediction_ood}, protein fitness prediction~\citep{paper-protein_disjoint_ddg}, and binding affinity prediction~\citep{paper-resolving_data_bias}. In some settings, a k-nearest-neighbor baseline can even rival more complex predictors~\citep{paper-similarity_target_prediction}. This illustrates why random splits can reward memorization of local similarity rather than generalization to independent samples.

Consequently, benchmark performance may not reflect performance on production data. This discrepancy may contribute to the generalization gaps reported in several areas of biochemical machine learning, including protein--ligand affinity modeling, TCR--peptide binding, pMHC prediction, and DNA-encoded-library screening~\citep{paper-cordial_generalization, paper-tcr_hard_split, paper-geometric_pmhc_generalization, paper-del_iver_ood}. Relation-aware splitting methods address this issue by keeping related samples within the same split (train or test), thereby preventing leakage. However, their use is currently limited by modality-specific assumptions, by a computational cost that prevents application to moderate to large datasets, and, for many, by a similarity threshold defining dependence for which no selection method exists, leaving users to fall back on conventional values that may not suit the task at hand \citep{paper-graphpart, paper-lohi}.

In this work, we introduce \NAME (\textbf{Re}lation-based \textbf{La}tent \textbf{G}rouping), a generic and modality-agnostic framework for relation-aware dataset splitting. We formalize a two-level latent-variable model in which observed samples are conditioned on group-specific latent variables, such that samples sharing the same latent variable form a group of related samples. From this model, we derive a practical algorithm that infers these groups from observed data. Given a modality-specific distance function, \NAME constructs an approximate proximity graph, identifies communities as estimates of these groups, and treats each community as an indivisible unit during random partitioning. By assigning entire communities to a single subset, \NAME prevents related samples from occurring on both sides of a subset boundary, making both subsets independent under the latent-variable model.

Using an approximate proximity graph, \NAME runs the most expensive step of the algorithm in $O(n\log n)$ time rather than the $O(n^2)$ cost of previous approaches that rely on pairwise comparisons \citep{paper-datasail, paper-hestia}. Across molecular and protein datasets, it yields minimal leakage comparable to existing methods \citep{paper-datasail, paper-hestia} while scaling more favorably for larger datasets, enabling relation-aware splits at scale.

The proximity graph depends on a distance threshold setting the resolution at which samples are considered related. We propose a method to select the most promising threshold from unlabeled production data as the resolution at which production and training data are most similar at the community level. Without relational structure, \NAME reduces to random splitting, thereby generalizing it. Conversely, when no threshold aligns them, the production data are detected as out of distribution.

The community structure additionally provides an approximate effective dataset size, $N_{\mathrm{eff}}$, defined by the number of communities. Across our experiments, $N_{\mathrm{eff}}$ is associated with generalization performance than the raw number of samples. This suggests that $N_{\mathrm{eff}}$ could support more efficient dataset scaling by prioritizing experiments expected to yield new communities.

Our contributions are as follows:
\begin{itemize}
    \item We formalize a modality-independent generative process for datasets containing related samples, providing a common foundation for relation-aware splitting across domains.
    \item We derive \NAME, an approach for partitioning any datasets generated under this process at scale by avoiding the quadratic cost of pairwise comparisons.
    \item We introduce a label-free threshold-selection procedure that adapts the splitting resolution to the intended production data.
    \item We show that \NAME's community structure provides a useful estimate of effective dataset diversity with potential to guide more efficient dataset scaling.
\end{itemize}

Although we focus on biochemical datasets, where sample relatedness is particularly salient and measurable, the underlying formalism applies to any domain with related samples that can be described by the proposed generative process, such as social networks or recommendation systems.

\section{Related Work}
Relation-aware splitting methods can be broadly grouped into optimization-based, clustering-based, and graph-based approaches. Optimization-based methods define an explicit criterion for a desirable split. For molecular data, SIMPD uses a multi-objective genetic algorithm to construct splits that mimic temporal drift in medicinal chemistry~\citep{paper-simpd}, whereas MOOD selects, among existing splitting protocols, the one whose train--test distance distributions best match a target deployment setting \citep{paper-mood}. DataSAIL formulates the problem more generally as a combinatorial optimization problem that minimizes cross-subset similarity~\citep{paper-datasail}.

Clustering-based approaches first cluster similar samples in the Euclidean space and assign entire clusters to subsets. For molecules, hierarchical clustering splits apply single-linkage or HDBSCAN clustering to ECFP4 fingerprints~\citep{paper-hierarchical_clustering_split}; alternative work instead proposes UMAP-based clustering~\citep{paper-scaffold_split_overestimate}. For biological sequences, SpanSeq combines alignment-free $k$-mer distances with DBSCAN clustering and constrained cluster assignment~\citep{paper-spanseq}, while Pfam-based splits assign protein families to folds~\citep{paper-pfam_cluster_generalization}.

Graph-based methods partition an explicit similarity graph. Lo-Hi formulates molecular dataset split as a balanced vertex $k$-cut problem~\citep{paper-lohi}. For biological sequences, dataset are partitioned along connected components of a thresholded similarity graph in binding affinity prediction~\citep{paper-latent_biases_binding_affinity} and by AutoPeptideML~\citep{paper-autopeptideml}. QMAP further subdivides large connected components through Leiden community detection~\citep{paper-qmap}. GraphPart constructs a weighted sequence-similarity graph and iteratively reassigns or removes samples until no cross-subset edge exceeds a specified threshold~\citep{paper-graphpart}. Hestia generalizes the GraphPart framework beyond sequences to multiple data modalities and out-of-distribution prediction settings~\citep{paper-hestia}. Structural-interface similarity graphs have also been used to reveal leakage in protein--protein interaction benchmarks~\citep{paper-protein_interaction_leakage}. Graph-based methods rely on a similarity threshold defining which samples are connected, which is typically set by convention.

Among these, DataSAIL and Hestia are the only generic methods designed to operate across modalities, but they pursue different objectives. DataSAIL minimizes similarity between subsets \citep{paper-datasail}, whereas Hestia retrains a model across a threshold sweep to estimate performance under increasing train--test dissimilarity \citep{paper-hestia}, an informative but costly aggregate compared to a single split. \NAME instead starts from a modality-independent generative model of relatedness and produces, at scale, subsets that are independent under it, at a resolution inferred from unlabeled production data to align evaluation with the deployment setting.

\begin{figure}[t]
    \centering
    \begin{subfigure}[b]{0.49\textwidth}
        \centering
        \includegraphics[]{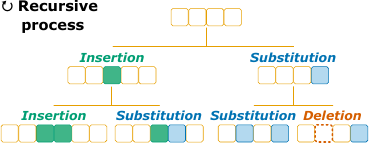}
        \caption{\textbf{Evolution:} sequences diverge from common ancestors via insertions, substitutions and deletions.}
        \label{fig:sub1}
    \end{subfigure}
    \hfill
    \begin{subfigure}[b]{0.49\textwidth}
        \centering
        \includegraphics[]{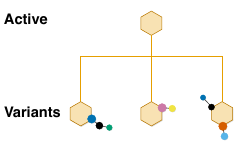}
        \caption{\textbf{Fragment based design:} drug hits are refined into close variants sharing a core scaffold.}
        \label{fig:sub2}
    \end{subfigure}
    \caption{Illustration of two generative processes that produce datasets containing related samples.}
    \label{fig:gen-process}
    \vspace{-.5\baselineskip}
\end{figure}
\section{Problem Formulation}
\begin{wrapfigure}{r}{0.35\linewidth}
    \vspace{-4\baselineskip}
    \centering
    \includegraphics[width=\linewidth]{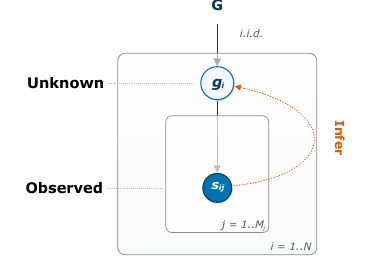}
    \caption{\textbf{Two-level latent-variable model.} Observed samples $s_{ij}$ are generated conditionally on group-specific latent variables $g_i$, which are drawn i.i.d. from a fixed population-level distribution $G$. The objective is to infer this group structure from the observed samples.}
    \label{fig:LVM}
    \vspace{-2\baselineskip}
\end{wrapfigure}

\ourparagraph{Generative process.}
\label{sec:generative-process}
Many datasets arise from generative processes in which common underlying sources produce multiple related observations. Biochemical data provide clear examples of this (Fig.~\ref{fig:gen-process}). Biological sequences, such as proteins, peptides, or genes, descend from ancestral sequences through evolution~\citep{paper-moelcular_evo_book}, such that sequences sharing a recent common ancestor tend to be close in edit distance (Fig.~\ref{fig:sub1})~\citep{paper-substitution_model}. Similarly, in bioactive molecule discovery, chemists synthesize local variants around active hits or combinatorial scaffolds, producing families of structurally related molecules (Fig.~\ref{fig:sub2})~\citep{paper-bioai_drug_discovery_perspective, paper-del_libraries}.
Although evolution and molecular screening operate on different objects, they induce the same dataset-level structure: samples are organized into groups of related observations rather than being independently and identically distributed.

\ourparagraph{A latent variable model of relatedness.}
We formalize the shared structure induced by these generative processes using a two-level latent-variable model (Fig.~\ref{fig:LVM}). Let $g_i$ denote an unobserved group-specific latent variable, drawn independently from a population-level distribution $G$. The $j$th sample from group $i$ is then drawn from a distribution conditioned on $g_i$:
\newcommand{\iidsim}{\overset{\smash{\scriptscriptstyle\text{i.i.d.}}}{\sim}}
$$
g_i \iidsim G,
\qquad
s_{ij} \mid g_i \sim p(s \mid g_i).
$$
Samples sharing the same $g_i$ are dependent when marginalizing over $g_i$, while samples from different groups are independent, so the i.i.d. assumption only holds at the group level. For instance, a group may correspond to a combinatorial molecular library, variants of a drug hit, a family or superfamily of proteins~\citep{paper-pfam}. The appropriate grouping thus depends on the resolution at which the problem is posed, which we make explicit through a resolution parameter $\tau$ and write $g_i \iidsim  G(\tau)$.

\section{Proposed Methods}
\subsection{The \NAME Algorithm}
\begin{figure}
    \centering
    \includegraphics[width=\textwidth]{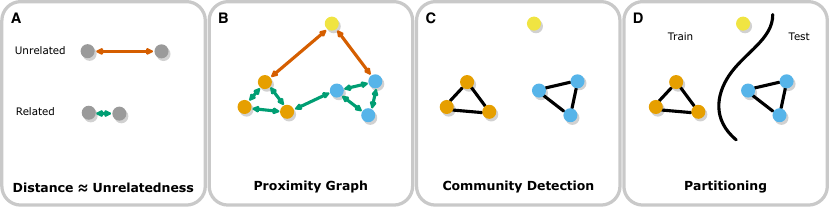}
    \caption{\textbf{\NAME algorithm overview.}
    \textbf{(A)} The algorithm requires a distance function that correlates with the probability that two samples are unrelated.
    \textbf{(B)} A proximity graph is built using this distance function.
    \textbf{(C)} Communities are identified, then graph edges that are unlikely to represent relationships are removed. Each community is an estimate of a group, i.e., a set of samples conditioned on the same latent variable $g_i$.
    \textbf{(D)} The dataset is partitioned along the community boundaries.}
    \label{fig:algo}
\end{figure}
The objective of \NAME is to infer, from the observed data, groups of related samples that are conditioned on the same latent variable $g_i$ (dotted orange arrows in Fig.~\ref{fig:LVM}). We refer to these inferred groups as \emph{communities}. Since groups are mutually independent under the model, each community is treated as an atomic unit: all of its samples must be assigned to the same split (e.g., training, validation, or test). A standard random split can then be applied over communities rather than over individual samples, preventing relations from crossing subset boundaries. It is achieved in four steps: defining a distance function, constructing an approximate proximity graph, identifying communities through community detection, and finally, partitioning (Fig.~\ref{fig:algo}).

\ourparagraph{Distance function.}
The method is modality-agnostic in that it requires only a distance function appropriate for the data. Consider two samples $s_{ij}$ and $s_{kl}$, belonging to groups $i$ and $k$, respectively. We require a distance $d$ for which proximity is informative of shared group membership:
$$
P\bigl(i \neq k \mid s_{ij}, s_{kl}\bigr)
= f\!\left(d(s_{ij}, s_{kl})\right),
$$
where $f$ is non-decreasing. 
Thus, larger distances correspond to a higher probability that two samples originate from different groups, and the smaller the distance, the more probable they belong in the same group~(Fig.~\ref{fig:algo}A). 

\ourparagraph{Approximate proximity graph.}
Given a distance threshold $\tau$, we define a proximity graph whose nodes are samples and whose edges connect pairs satisfying $d(s_{ij}, s_{kl}) < \tau$ (Fig.~\ref{fig:algo}B). Constructing this graph exactly requires $n(n-1)/2$ distance evaluations and therefore an $O(n^2)$ time complexity. This can be prohibitive for large datasets or expensive distance functions. We instead construct an approximate proximity graph using the Hierarchical Navigable Small World (HNSW) data structure~\citep{paper-hnsw}. HNSW organizes samples into exponentially sparser graph layers, allowing for approximate nearest-neighbor insertion and retrieval in $O(\log n)$ time. Standard HNSW is optimized for nearest-neighbor retrieval speed and therefore stores only a sparse subset of the pairwise relationships encountered during graph construction. However, many additional sample pairs are compared while searching for neighbors, and subsequently discarded from the HNSW structure. For proximity-graph construction, these evaluations remain informative: any evaluated pair satisfying $d(s_{ij},s_{kl})<\tau$ provides evidence of a candidate relation. We therefore augment the standard HNSW construction by retaining all evaluated pairs whose distance falls below $\tau$ as proximity-graph edges, rather than only those selected as HNSW edges. This yields a denser approximation of the proximity graph without requiring additional distance evaluations (Fig.~\ref{fig:algo}B).

\ourparagraph{Community detection.}
The distance function provides only a noisy proxy for group membership: samples from different groups can nevertheless fall within the distance threshold by chance, thereby inducing spurious proximity-graph edges. Under a null model in which proximity arises by chance between unrelated samples, an edge occurs with probability $p_0 = P(d(s_{ij},s_{kl}){\,<\,}\tau \mid i {\,\neq\,} k)$, independently of group membership. Consequently, such edges are uniformly dispersed across groups. In contrast, samples from the same group are more likely to lie within the threshold and therefore induce an excess of within-group edges. Thus, groups manifest as regions of the proximity graph whose internal connectivity exceeds that expected under the null model.

We estimate $p_0$ using a modality- and problem-specific reference procedure that removes genuine relatedness while preserving the marginal data distribution, for example, shuffling residues within sequences, sampling random valid molecules, or pairing images from distinct classes. This procedure yields the distance distribution of unrelated pairs from which $p_0$ can be estimated. We then apply the Leiden community-detection algorithm~\citep{paper-leiden} with the Constant Potts Model (CPM) objective~\citep{paper-CPM_objective}:
$
\sum_{c} \left[e_c - \gamma \textstyle\binom{n_c}{2}\right],
$
where $e_c$ and $n_c$ are the number of internal edges and nodes in community $c$, respectively, and $\gamma=p_0$ is the null edge probability. Maximizing this objective identifies communities whose internal edge density exceeds that expected under the null model~(Fig.~\ref{fig:algo}C). When a modality-specific null model is unavailable, communities can be obtained using a traditional Modularity objective~\citep{paper-leiden}.

\ourparagraph{Community-level partitioning.}
Finally, we treat each community as an atomic unit and apply a random train--test split over communities rather than individual samples. Consequently, all samples within a community are assigned to the same subset, resulting in independent subsets under the latent-variable model~(Fig.~\ref{fig:algo}D).

\subsection{Selecting the Distance Threshold}
\begin{wrapfigure}{r}{0.4\linewidth}
    \vspace{-4\baselineskip}
    \centering
    \includegraphics[width=0.4\textwidth]{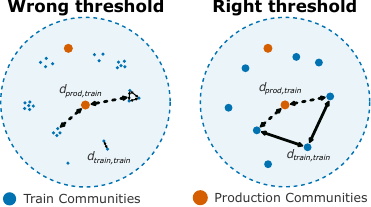}
    \caption{\textbf{Threshold selection intuition.} A wrong threshold makes train communities closer or further away from each other 
    than they are from production communities. The right threshold makes production communities integrate nicely in the community distribution.}
    \label{fig:threshold_algo}
    \vspace{-2\baselineskip}
\end{wrapfigure}
The chosen distance threshold $\tau$ has a large impact on the obtained communities.
For some data modalities, such as biological sequences, the latent-variable model is resolution-dependent: the same dataset can admit valid groupings at multiple levels of relatedness. In \NAME, this resolution is controlled by the distance threshold $\tau$ used to construct the proximity graph. Larger values of $\tau$ connect more samples and yield coarser groups, whereas smaller values produce finer groups. Since the true resolution is unobserved, finding the right threshold for the task at hand is challenging.

In the following, we propose an automated scheme to select the most promising threshold $\tau^\star$ using unlabeled production data, i.e., samples on which the trained model is expected to be deployed. Such samples are often available before labels are collected. For instance, in hit discovery, candidate molecules can be sampled from the screening library before any assay is run. The hypothesis is that the production samples are drawn from the same distribution as the training data. Here, training data refer to the full dataset to be split prior to partitioning. Under this hypothesis, the appropriate resolution is the one at which production and training samples are indistinguishable at the group level, as illustrated in \autoref{fig:threshold_algo}. Intuitively, $\tau^\star$ should make the distribution of distances among training communities match the distribution between production and training communities.

\newcommand{\Ctau}{\mathcal{C}^{\tau}}
We search for $\tau^\star$ by sweeping over candidate thresholds. To keep the sweep tractable, we build a single HNSW graph over the union of training and production samples and use its sparse base layer as a candidate graph. For each candidate threshold~$\tau$, we retain edges shorter than $\tau$, estimate the null model, and apply Leiden community detection with the CPM objective. This yields three sets of communities: $\Ctau_{\mathrm{train}}$ denotes the set of \emph{pure-training} communities (containing only training samples); $\Ctau_{\mathrm{prod}}$ denotes \emph{pure-production} communities (only production samples); and $\Ctau_{\mathrm{hybrid}}$ denotes \emph{hybrid} communities (samples from both sets).
For each threshold, we compare how far pure-training and pure-production communities lie from the training data. We measure the distance between two communities $A$ and $B$ by their single-linkage distance, i.e., the smallest distance between any sample of one and any sample of the other, $d_{\mathrm{SL}}(A, B) = \min_{a \in A,\, b \in B} d(a, b)$. 
We then consider the empirical cumulative distribution functions of single-linkage distances, from every pure-training and from every pure-production community to every other training-containing community:
\begin{align}
F^{\tau}_{\mathrm{train}\to\mathrm{train}}(x) &= \tfrac1{n_1} \left|\big\{ d_{\mathrm{SL}}(A, B) \leq x \;:\; A \in \Ctau_{\mathrm{train}},\; B \in \Ctau_{\mathrm{train}}\cup \Ctau_{\mathrm{hybrid}}, A \neq B \big\}\right|, \\
F^{\tau}_{\mathrm{prod}\to\mathrm{train}} (x) &= \tfrac1{n_2}\left|\big\{ d_{\mathrm{SL}}(A, B)\leq x \;:\; A \in \Ctau_{\mathrm{prod}},\; B \in \Ctau_{\mathrm{train}}\cup \Ctau_{\mathrm{hybrid}} \big\}\right|,
\end{align}
with $n_1$ and $n_2$ chosen such that $F^{\tau}_{\mathrm{train}\to\mathrm{train}}(\infty) {=} F^{\tau}_{\mathrm{prod}\to\mathrm{train}}(\infty) {=} 1$.
We compare these distributions using the Kolmogorov-Smirnov statistic \citep{paper-KS-test} 
${\rm KS}(F_1, F_2) = \max_x|F_1(x){-}F_2(x)|$ and select the $\tau$ that minimizes it:
$
\tau^\star = \arg\min_{\tau} \operatorname{KS}\!\left(
F_{\mathrm{train}\rightarrow\mathrm{train}}^{\tau},
F_{\mathrm{prod}\rightarrow\mathrm{train}}^{\tau}
\right).
$

The procedure returns $\tau^\star$ together with the minimal KS statistic, which measures how closely production and training data can be aligned. A value near 0 indicates that the hypothesis holds at resolution $\tau^\star$ while a value near 1 indicates that no threshold can align the two distributions, signaling that the production data are out of distribution. The full procedure is given in Algorithm~\ref{alg:threshold}.

\section{Evaluation}
\begin{figure}[t]
    \centering
    \begin{subfigure}[t]{0.32\textwidth}
        \centering
        \includegraphics[width=\linewidth]{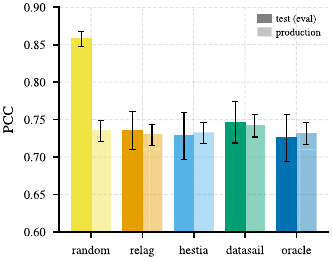}
        \caption{\!Mean Pearson correlation (PCC) and std between $\hat{y}$ and $y$ on the test split and on production data. 
        }
        \label{fig:synthetic_test_vs_prod}
    \end{subfigure}\hfill
    \begin{subfigure}[t]{0.32\textwidth}
        \centering
        \includegraphics[width=\linewidth]{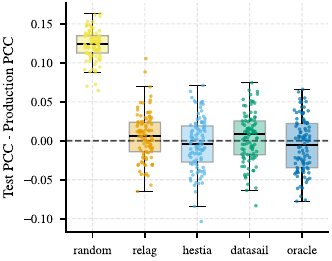}
        \caption{Per-repeat difference between test and production PCC. $0$ indicates a faithful evaluation.}
        \label{fig:synthetic_eval_gap}
    \end{subfigure}\hfill
    \begin{subfigure}[t]{0.32\textwidth}
        \centering
        \includegraphics[width=\linewidth]{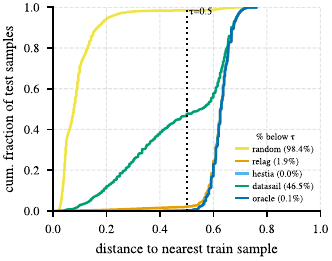}
        \caption{ECDFs of the distance ($1 -$ global identity) from each test sample to its nearest training sample.
        }
        \label{fig:synthetic_leakage}
    \end{subfigure}
    \caption{\textbf{Relation-aware splits yield evaluations that match production performance.}
    In each of 100 repeats, we sample independently two synthetic peptide datasets (15{,}000 sequences, 1{,}000 families each) under the latent-variable model: an annotated dataset to split and a production dataset. 
    }
    \label{fig:synthetic}
    \vspace{-.5\baselineskip}
\end{figure}
\subsection{Synthetic validation}
We compare \NAME with DataSAIL~\citep{paper-datasail} and Hestia~\citep{paper-hestia}, the two existing relation-aware splitting methods designed to operate across data modalities, and with a random-splitting baseline. Assessing whether a split yields a faithful estimate of generalization requires a labeled production dataset, which benchmark datasets lack. We therefore first evaluate on synthetic data. As reference point, we include an oracle that splits along true groups.

\ourparagraph{Setup.} We instantiate the latent-variable model with a peptide generative process in which each family descends from an independently sampled ancestral sequence through recursive mutations (Appendix). Labels are given by a deterministic function of the sequence corrupted by Gaussian noise. For each of 100 repeats, we draw two independent datasets from this process: an annotated dataset to be partitioned and a production dataset. The threshold selection procedure yields $\tau^\star = 0.5$ (\autoref{fig:synthetic_threshold}), which we use for all threshold-based methods and at which \NAME communities recover the ground-truth families almost exactly (the average Adjusted Rand Index is $0.991 \pm 0.006$).

\ourparagraph{Test versus production.} For each split, we encode peptides using ESM Cambrian~\citep{paper-esmc}, train a two-layer MLP prediction head on the training subset, and measure its Pearson correlation (PCC) between predictions $\hat{y}$ and targets $y$ on both the test subset and the production dataset. Random splitting overestimates production performance substantially with a test PCC of $0.86$ compared to $0.74$ in production (\autoref{fig:synthetic_test_vs_prod}). In contrast, the test performance of every relation-aware method closely matches its production performance. Per repeat, the difference between test and production PCC is centered on zero for \NAME ($+0.005$), Hestia ($-0.004$), DataSAIL ($+0.004$), and the oracle ($-0.006$) (\autoref{fig:synthetic_eval_gap}).

\ourparagraph{Train--test distance.} The source of this overestimation is visible in the distance from each test sample to its nearest training sample (\autoref{fig:synthetic_leakage}). Under random splitting, $98.4\%$ of test samples lie within $\tau^\star$ of a training sample, i.e., nearly every test sample has a relative in the training set. A model can thus achieve high test performance by exploiting similarity to training samples rather than learning the underlying sequence--property relationship, an advantage that vanishes on unseen families~\citep{paper-similarity_target_prediction}.  \NAME ($1.9\%$) and Hestia ($0.0\%$) nearly match the oracle ($0.1\%$), whereas DataSAIL leaves $46.5\%$ of test samples within $\tau$ of the training set. Because DataSAIL minimizes overall train--test similarity rather than enforcing a specific threshold, it does not guarantee that individual test samples lack close relatives in training.
Together, these results show that random splitting overestimates generalization when samples are related, whereas relation-aware splits yield test performance that reflects production performance. \NAME performs on par with existing relation-aware methods.

\begin{wrapfigure}{r}{0.4\linewidth}
    \centering
    \vspace{-1\baselineskip}
    \includegraphics[width=0.4\textwidth]{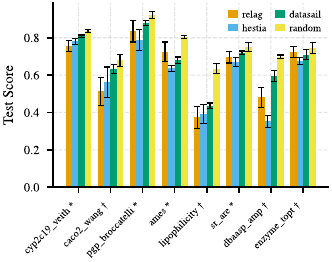}
    \caption{\textbf{Performances on real data.}  Downstream MLP test performance of dataset-splitting strategies across eight datasets (six molecular and two protein datasets). Results are averaged over 10 split seeds, with error bars denoting standard deviations. We report AUROC for classification tasks and Pearson correlation coefficient for regression tasks.}
    \label{fig:perf-combined-mlp}
\end{wrapfigure}

\subsection{Real data}
We next evaluate the same methods on real data, comprising six molecular and two protein datasets. Although these datasets lack production data, we examine whether they exhibit the same train--test distance and test performance patterns as the synthetic data.

\ourparagraph{Train--test distance.}
For a controlled comparison on real data, we use a common distance threshold for all datasets within each modality: $0.5$ for proteins and $0.6$ for molecules. As on synthetic data, random splitting leaves most test samples within the threshold of a training sample (79--82\%), whereas \NAME yields residual similarity comparable to Hestia and lower than DataSAIL across both modalities (\autoref{fig:leakage-ecdf}). For proteins, \NAME places only $1\%$ of test samples below the threshold, compared to $10\%$ for Hestia (\autoref{fig:leakage-protein}). For molecules, Hestia places no test samples below the threshold, compared to $23\%$ for \NAME (\autoref{fig:leakage-molecule}). This difference reflects distinct objectives. Hestia removes all train--test similarity below a chosen threshold, explicitly constructing an out-of-distribution test set. \NAME instead targets statistical independence under the latent-variable model, allowing residual similarity between samples of different groups that fall within the threshold by chance.

\ourparagraph{Test performance.}
We follow the same protocol as for synthetic data, encoding proteins with ESM Cambrian~\citep{paper-esmc} and molecules with ChemBERTa~\citep{paper-chemberta}, and repeat each experiment over 10 split seeds. Random splitting yields higher test performance than every relation-aware method, consistent with the overestimation observed on synthetic data, whereas \NAME, Hestia, and DataSAIL yield comparable performance (\autoref{fig:perf-combined-mlp}). Although production performance cannot be measured on these datasets, relation-aware methods agree as they do on synthetic data, where their test estimates matched production.

\begin{figure}[t]
    \centering
    \begin{subfigure}{0.32\textwidth}
        \centering
        \includegraphics[width=\textwidth]{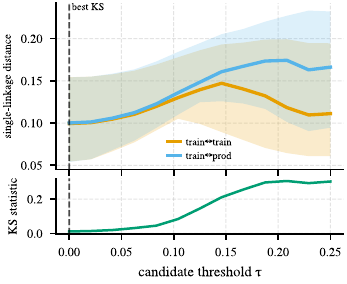}
        \caption{i.i.d.: MNIST.}
        \label{fig:threshold_MNIST}
    \end{subfigure}
    \hfill
    \begin{subfigure}{0.32\textwidth}
        \centering
        \includegraphics[width=\textwidth]{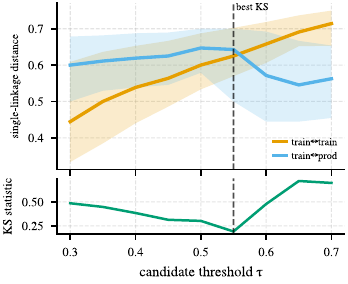}
        \caption{Related: DBAASP (training) vs. PeptideAtlas (Production).}
        \label{fig:threshold_dbaasp}
    \end{subfigure}
    \hfill
    \begin{subfigure}{0.32\textwidth}
        \centering
        \includegraphics[width=\textwidth]{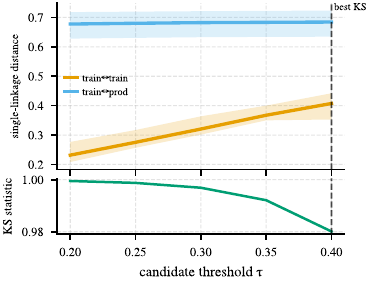}
        \caption{OOD: BELKA training set vs. non-triazine test subset.}
        \label{fig:threshold_belka}
    \end{subfigure}
    \caption{\textbf{Threshold selection characterizes the relatedness regime.} Top: single-linkage distances from pure-training and pure-production communities to training communities (15th–85th percentile shaded). Bottom: KS statistic.}
    \label{fig:4}
    \vspace{-0.8\baselineskip}
\end{figure}

\ourparagraph{Scalability.}
Beyond evaluation quality, the computational cost of existing relation-aware methods limits their application to large datasets.
We therefore compare the runtime of \NAME and Hestia, the two fastest of the studied methods, on increasingly large subsets of PeptideAtlas~\citep{paper-peptide_altas}. Runs exceeding one day are terminated, and their runtimes are extrapolated from fitted scaling curves. 
\begin{wrapfigure}{r}{0.4\linewidth}
    \centering
    \includegraphics[width=0.4\textwidth]{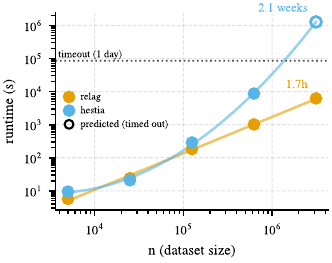}
    \caption{\textbf{Scaling on PeptideAtlas subsets.} Runtime as a function of dataset size for \NAME and Hestia, the two fastest compared methods. Runs were limited to one day; values above the dashed line are extrapolated. Log--log fits indicate approximately linear scaling for \NAME and quadratic scaling for Hestia ($R^2=0.998$ for both).}
    \vspace{-3\baselineskip}
    \label{fig:scaling}
\end{wrapfigure}
As shown in \autoref{fig:scaling}, the methods have comparable runtime for datasets containing tens of thousands of samples, but \NAME scales more favorably as dataset size increases. At $n=3.125$M, Hestia exceeds the one-day time budget, with an extrapolated runtime of 2.1 weeks, whereas \NAME completes in 1.7 hours, an estimated $207\times$ speedup that enables relation-aware splitting at previously impractical dataset scales. We observe a similar trend on the molecule modality (Appendix~\autoref{fig:scaling_belka}).

\subsection{Threshold}
We next evaluate the threshold-selection procedure across three relatedness regimes. For data without detectable relational structure, we expect the selected threshold to be close to zero since all groups are singletons. For relational data, the procedure should select a non-zero threshold corresponding to a meaningful grouping resolution. Finally, when production data are out-of-distribution (OOD), no threshold should align the train--train and production--train distance distributions.

We consider three datasets illustrating these regimes (Fig.~\ref{fig:4}). MNIST~\citep{paper-mnist} provides an approximately i.i.d.\ setting. By treating the test set as production data, the KS statistic is minimized at $\tau=0$ (Fig.~\ref{fig:threshold_MNIST}), consistent with the absence of relational structure. DBAASP~\citep{paper-dbaasp}, a dataset of antimicrobial peptides with substantial sequence relatedness, represents the relational regime. Using the PeptideAtlas database~\citep{paper-peptide_altas} as production data, the threshold converges at $\tau=0.55$ (Fig.~\ref{fig:threshold_dbaasp}), indicating a recoverable relational grouping. Finally, BELKA~\citep{paper-belka}, a DNA-encoded-library screening benchmark released through a NeurIPS 2024 Kaggle competition, tests the OOD regime: a post-competition analysis found that none of the $\sim$2,000 participating teams outperformed random predictions on the designated OOD test subset~\citep{paper-del_iver_ood}. Its training molecules all contain a triazine core absent from all molecules of that test subset, and although training samples become densely connected by $\tau=0.4$, no threshold aligns the two distance distributions (minimum KS statistic of $0.98$; Fig.~\ref{fig:threshold_belka}), consistent with the production samples being OOD relative to training.

These regimes illustrate that \NAME can be seen as generalizing random splitting along the relatedness axis. When no relational structure is detected, each sample forms its own group, and group-level random splitting recovers conventional random splitting. When groups are present, \NAME partitions along their inferred boundaries. At the opposite extreme, when all samples belong to a single group, no non-leaking split is possible, and \NAME instead signals that the training and production datasets are mutually out-of-distribution.

\subsection{Free diversity estimate}
\label{sec:neff}
Our formalism and algorithm provide a natural estimate of the effective dataset size: $N_{\mathrm{eff}}$, corresponding to the number of communities. We evaluate whether $N_{\mathrm{eff}}$ is more informative of downstream generalization than the raw dataset size $N$ across our benchmark datasets.

\begin{figure}
    \centering
    \begin{subfigure}{0.48\textwidth}
        \centering
        \includegraphics[width=.9\textwidth]{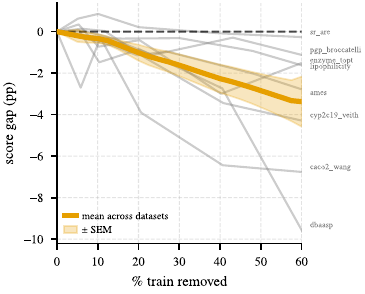}
        \caption{
        \textbf{Removing random samples vs communities} The test-score difference between removing samples uniformly at random and complete communities is shown as a function of the fraction of training data removed. Thin lines denote individual datasets; the thick line and shaded region denote the mean and SEM, respectively. Negative values indicate that removing entire communities produces a larger performance decrease.
        }
        \label{fig:data-starvation_lines}
    \end{subfigure}
    \hfill
    \begin{subfigure}{0.48\textwidth}
        \centering
        \includegraphics[width=.9\textwidth]{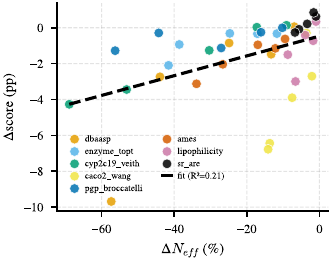}
        \caption{\textbf{Effective dataset size.} Each point shows the difference in test performance when dropping entire train communities compared to dropping train samples at random, ensuring the training set sizes are equal across both conditions. The x-axis shows the resulting difference in $N_{\mathrm{eff}}$, the number of train communities, and the y-axis the difference in test score (community - random). Larger losses in $N_{\mathrm{eff}}$ are associated with larger performance drops (r = 0.46).}
        \label{fig:data-starvation_corr}
    \end{subfigure}
    \caption{
        Community structure captures an aspect of training-set diversity that is not reflected by the raw sample count and is associated with generalization performance.
    }
    \vspace{-1\baselineskip}
    \label{fig:data-starvation}
\end{figure}

Since most analyzed benchmarks have no production data, we choose for each dataset a threshold at which community sizes follow a realistic heavy-tailed distribution as too small thresholds fragment the data into mostly small communities. We then remove either complete communities or an equal number of randomly selected samples from the training set. Removing entire communities produces a larger performance decrease than removing the same number of samples at random (Fig.~\ref{fig:data-starvation_lines}), indicating that the loss of entire communities is more harmful to downstream performance than a comparable reduction in raw sample count. Across datasets, the performance gap is correlated with reductions in $N_{\mathrm{eff}}$ (Fig.~\ref{fig:data-starvation_corr}). This pattern is consistent with within-group redundancy limiting the contribution of related samples to generalization~\citep{paper-n_eff_og}.

The BELKA dataset provides an illustrative limiting case. As discussed previously, because the production set is OOD, the training set's proximity graph collapses to a complete graph (i.e., every sample is connected to every other sample) before the two distance distributions can align. This yields $N_{\mathrm{eff}}=1$ and indicates that the training data alone do not provide independent groups from which to generalize to other groups such as the test one.  Doing so would require increasing the number of groups, thus diversity.

More broadly, these results suggest a strategy for more efficient dataset scaling: prioritize the acquisition and annotation of samples that form new communities, rather than samples that primarily increase the size of existing ones.

\section{Conclusion}
We introduced a relation-aware splitting framework derived from a latent-variable model of sample relatedness. By inferring groups through an approximate proximity graph and community detection, \NAME constructs train--test splits that preserve inferred group boundaries. Across molecular and protein datasets, it provides leakage-aware evaluations comparable to existing generic methods while enabling substantially larger-scale splits. A label-free threshold-selection procedure further adapts the splitting resolution to production data, recovering random splitting when no relational structure exists and flagging out-of-distribution production data. Its communities additionally provide an effective dataset size, $N_{\mathrm{eff}}$, which may support diversity-aware dataset scaling.

Although \NAME is fast relative to other relation-aware methods, it remains slower than random splitting and is therefore unnecessary when samples are known to be independent. The method also requires a meaningful distance function, which may be non-trivial to define for some modalities. Threshold selection further requires unlabeled production data, which are not always available. Finally, without a modality-specific null model, \NAME falls back to the less interpretable modularity objective. Future work will further explore the generality of \NAME by applying its framework beyond biochemical data.

\section{AI use statement}
In this work, we used generative AI tools to formulate mathematical equations, design and provide feedback on research experiments. We have not used generative AI tools to help develop theoretical models or conceptual frameworks, formulate mathematical claims, provide critical ingredients for proving mathematical claims, propose or refine hypotheses, implement methods, assist with translation, clean and reformat dataset, support qualitative and thematic data analysis, interpret results, and generate synthetic data sets and assist in the writing of proofs, are not applicable to this work. Additionally, we used generative AI tools for create or modify scientific figures or images, create or edit software code, draft parts of a research paper, brainstorming, sourcing/searching for information, edit a research paper to improve readability, identify relevant literature and propose a title and keywords for a research paper. We have reviewed all AI-assisted work. For each paper identified, we verified that it supported the cited claims and conducted traditional literature searches to minimize missed publications. For software package development, we reviewed and tested all AI-generated code to ensure it functioned as expected and met our quality standards. For research experiments, we verified that the code accurately implemented the methodology and critically assessed the robustness and scientific validity of the experimental design. We critically evaluated any scientific feedback provided by generative AI models rather than accepting it without verification. Mathematical equations were independently verified by at least two co-authors to ensure their mathematical soundness. For figure generation, AI was used to assist in writing the code, while we independently verified that the figures accurately represented the underlying data. Finally, any text reformulated by AI was reviewed by the authors for accuracy and fidelity to the original content. We take responsibility for the final content of this work, including text, claims or artifacts produced with the aid of generative AI.

\section{Reproducibility statement}
We provide the complete implementation of \NAME as an open-source Rust library with Python bindings available at: \url{https://github.com/anthol42/relag/tree/main}
Additionally, all code needed to reproduce our results is available at \url{https://github.com/anthol42/relag-paper}. The code automatically downloads and preprocesses the datasets, runs all experiments, and generates every figure in the paper. The repository README lists the scripts and commands for reproducing each experiment and figure.

\section{Acknowledgments}
Lavertu A. acknowledges funding from the Fonds de recherche du Québec through a Master's Research Scholarship (\href{https://doi.org/10.69777/369917}{doi: \textcolor{blue}{10.69777/369917}}) and NSERC's Canada Graduate Research Scholarship -- Master's program.
Côté J. acknowledges funding from the Fonds de recherche du Québec through a Master's Research Scholarship (\href{https://doi.org/10.69777/360097}{doi: \textcolor{blue}{10.69777/360097}}) and NSERC's Canada Graduate Research Scholarship -- Master's program.
Gobeil S. is supported by an NSERC Discovery Grant (RGPIN-2023-03546), an NSERC Discovery Launch Supplement (DGECR-2023-00091) and a CBRF and Biosciences Research Infrastructure Fund (BRIF) award for the ``Biologics RAMP-UP: Biologics Rapid Actuation of Mass Production Under Pandemic conditions'' project (CBRF2-2023-00176).
Corbeil J. acknowledges the support of the Canadian Biomanufacturing Research Fund (CBRF) and, specifically, the PandemicStopAI project: an accelerated response to bacterial pandemics. He also acknowledges the funds for the CQDM (Quebec Consortium for Drug Discovery).
I. Prémont-Schwarz is grateful to be supported by the CFREF IVADO Professor grant CFREF-2022-00051.
Germain P. is supported by by the NSERC Discovery grant RGPIN-2020-07223.
\bibliography{iclr2027_conference}
\bibliographystyle{iclr2027_conference}

\appendix
\section{Appendix}
\subsection{Additional Methods}
\subsubsection{Null Model}

Estimating the null edge probability,
$$
p_0(\tau)
=
P\!\left(d(s_{ij},s_{kl})<\tau \mid i \neq k\right),
$$
corresponding to the probability that two unrelated samples fall within distance $\tau$, is modality- and problem-specific. We therefore estimate the null distance distribution using Monte Carlo procedures designed to remove relational structure while preserving the marginal characteristics of the data.

For protein sequences, we sample $10$ million random pairs from PeptideAtlas~\citep{paper-peptide_altas}. For each sequence, we independently shuffle its amino acids before computing pairwise distances. This procedure destroys sequence homology while preserving sequence length and amino-acid composition, thereby retaining the marginal sequence distribution while removing relational structures. For molecules, directly shuffling molecular fingerprints would destroy chemical structure and produce distances substantially larger than those observed between real molecules, yielding an unrealistically small estimate of $p_0$. Instead, we generate $100,000$ chemically valid molecules using randomly generated SELFIES~\citep{paper-selfies}, with molecular sizes sampled from $\mathcal{N}(50,5)$, and compute distances for $10$ million random pairs. For MNIST, we construct the null distribution from a subset of ImageNet~\citep{paper-imagenet} by downscaling images to the MNIST resolution and converting them to grayscale. $10$ million random pairs are sampled from distinct ImageNet classes.

The probability $p_0(\tau)$ may correspond to a rare event, for which direct Monte Carlo estimation can be unreliable when few or no sample pairs satisfy $d<\tau$. We therefore extrapolate the lower $1\%$ tail by fitting a Generalized Pareto Distribution (GPD) to the smallest $1\%$ of sampled distances and estimate $p_0(\tau)$ from the fitted distribution. The resulting estimate is used as the CPM resolution parameter, i.e.,
$$
\gamma = p_0(\tau).
$$

\subsubsection{Data Preparation}

\paragraph{Datasets.}
We evaluate \NAME on eight real datasets, including six molecular datasets and two sequence-based datasets (Table~\ref{tab:datasets}). For all datasets, we discard the predefined train--test splits and re-split the pooled data from scratch.

The molecular datasets include five single-task datasets from the Therapeutics Data Commons (TDC)~\citep{paper-tdc}: \texttt{cyp2c19\_veith}, which measures inhibition of the CYP2C19 enzyme~\citep{paper-cyp2c19_veith}, \texttt{caco2\_wang}, which measures Caco-2 permeability~\citep{paper-caco2_wang}, \texttt{pgp\_broccatelli}, which measures P-glycoprotein inhibition~\citep{paper-pgp_broccatelli}, \texttt{ames}, which measures Ames mutagenicity~\citep{paper-ames}, and \texttt{lipophilicity}, which measures AstraZeneca logD~\citep{paper-lipophilicity}. We additionally use the SR-ARE task from the multi-task Tox21 dataset~\citep{paper-tox21}. For sequence-based datasets, we use DBAASP antimicrobial peptides~\citep{paper-dbaasp}, retaining only sequences composed of canonical L-amino acids with an \emph{E.~coli} MIC label, with labels expressed as $\log_{10}$ MIC obtained from the QMAP benchmark package \citep{paper-qmap}. We also use the enzyme optimal-temperature regression dataset~\citep{paper-enzyme_topt}.

\paragraph{Synthetic dataset.}
Additionally, we generate synthetic peptide datasets that follow the two-level latent-variable model, where each group is a family of sequences descending from a common ancestor. Each dataset contains $15{,}000$ sequences from $1{,}000$ families. For each family, we generate a random ancestral sequence: its length is drawn from the empirical length distribution of PeptideAtlas sequences of 20 to 50 residues~\citep{paper-peptide_altas}, and each residue is drawn independently from the amino-acid frequencies of PeptideAtlas. Families are then grown by a branching process: starting from the ancestor, each sequence produces a Poisson-distributed number of descendants (mean $1.6$), each obtained by applying $1 + \mathrm{Poisson}(2)$ random edits to its parent. An edit is a substitution ($85\%$), an insertion ($7.5\%$), or a deletion ($7.5\%$) at a uniformly sampled position, with substitutions drawn from the BLOSUM62 conditional distribution~\citep{paper-blosum} and insertions from the background frequencies. The accumulated number of edits from the ancestor is capped at $10$, and growth proceeds breadth-first until the target dataset size is reached. The annotated and production datasets are generated with different random seeds and therefore share no family.

The target is a non-linear function of three physicochemical descriptors computed with modlAMP~\citep{paper-modlAMP}: the net charge at pH 7, the maximal Eisenberg hydrophobic moment over 11-residue windows~\citep{paper-hydrophobic}, and the Boman index~\citep{paper-boman_index}. Let $z_c$, $z_m$ and $z_b$ denote the standardized charge, log hydrophobic moment and Boman index, and $s(x) = \tanh(0.7x)/0.7$ a soft saturation. The target is
$$
    y = f(x) + \varepsilon,
$$
where
$$
    f(x) = s(z_m) + 0.8\, s(z_c) + s(z_m)\, s(z_c) - s(z_m)\, s(z_b) + 0.8 \sin(1.5\, z_b).
$$
$f$ is standardized to unit variance and $\varepsilon \sim \mathcal{N}(0, 0.05^2)$. The standardization constants are fixed once, so annotated and production datasets share the same function. Since $y$ depends only on the sequence, any systematic  difference between test and production performance is attributable to the split.

\paragraph{Filtering.}
Filtering is performed once before splitting so that all methods and downstream models use identical sample indices. Molecular samples with missing SMILES or labels, or with unparsable SMILES, are removed. Protein samples with missing or non-finite labels are removed. We retain sequences composed of the 20 standard amino acids, along with \texttt{X} in \texttt{enzyme\_topt}, which is supported by ESM Cambrian. For PeptideAtlas~\citep{paper-peptide_altas}, which is used only for null-model estimation and scalability experiments, we merge all build FASTA files, deduplicate sequences, and retain sequences of length at most 100. For BELKA, we remove duplicates and extract the non-triazine subset of the test set as the OOD dataset.

\paragraph{Representations.}
For \NAME, molecules are represented using 2048-bit Morgan fingerprints with radius 2~\citep{paper-ecfp}, computed with RDKit. Protein sequences are represented directly as amino-acid strings. For downstream prediction, molecules are encoded using ChemBERTa (\texttt{seyonec/ChemBERTa-zinc-base-v1})~\citep{paper-chemberta}, using the \texttt{[CLS]} representation from the final hidden layer. Proteins are encoded using ESM Cambrian 300M~\citep{paper-esmc}, using mean-pooled final-layer residue embeddings. The encoders are frozen, and only the prediction head is trained. Distances used are $1 -$ Tanimoto similarity on Morgan fingerprints for molecules, $1 -$ global alignment identity~\citep{paper-global_alignment} for protein and peptide sequences, ProtSpaM for UniRef50 (A.3.2)~\citep{paper-protspam}, and $1 -$ cosine similarity for MNIST.

\begin{table}[h]
\centering
\small
\caption{Datasets used in the experiments. $N$ is the number of samples after filtering.}
\label{tab:datasets}
\begin{tabular}{lccccc}
\toprule
Dataset & Modality & $N$ & Task & Metric & Encoder \\
\midrule
\texttt{cyp2c19\_veith}   & molecule & 12{,}665 & classification & AUROC & ChemBERTa \\
\texttt{ames}             & molecule & 7{,}278  & classification & AUROC & ChemBERTa \\
\texttt{sr\_are}          & molecule & 5{,}832  & classification & AUROC & ChemBERTa \\
\texttt{lipophilicity}    & molecule & 4{,}200  & regression     & PCC   & ChemBERTa \\
\texttt{pgp\_broccatelli} & molecule & 1{,}218  & classification & AUROC & ChemBERTa \\
\texttt{caco2\_wang}      & molecule & 910      & regression     & PCC   & ChemBERTa \\
\texttt{dbaasp}      & peptide  & 10{,}664 & regression     & PCC   & ESM-C 300M \\
\texttt{enzyme\_topt}     & protein  & 1{,}886  & regression     & PCC   & ESM-C 300M \\
\bottomrule
\end{tabular}
\end{table}

\subsubsection{Split Comparison}

\paragraph{Protocol.}
All splitting methods operate on the same data, representations, and use the same MLP architecture. The only difference between methods is how the train--test split is constructed. We allocate 80\% of the samples to the training and validation set and 20\% to the test set. Of the 80\%, 10\% is randomly held out as a validation set for early stopping. We evaluate regression tasks using the Pearson correlation coefficient between $\hat{y}$ and $y$ (PCC) and classification tasks using AUROC. Each experiment is repeated over 10 random seeds (1--10) for each method and dataset.

\paragraph{Threshold.}
We use fixed thresholds within each modality to ensure a controlled comparison across splitting methods. Because the benchmark datasets contain no explicit production samples, the threshold-selection procedure cannot be applied. We therefore use a fixed threshold per modality: $\tau=0.60$ for molecular datasets and $\tau=0.50$ for protein datasets with \NAME. Equivalently, the corresponding thresholds for Hestia are 0.40 Tanimoto similarity for molecules and 50\% sequence identity for proteins. These thresholds are kept fixed across datasets within each modality to ensure a fair comparison.

\paragraph{Prediction head.}
For downstream evaluation, we use a two-layer MLP on top of the frozen representations. Given an encoder output of dimension $d$, the architecture is
\[
\texttt{Linear}(d,2d) \rightarrow \texttt{ReLU} \rightarrow \texttt{Linear}(2d,1).
\]
The model is optimized with Adam using a learning rate of $10^{-3}$ and a batch size of 64 for up to 300 epochs. Early stopping is performed using the validation loss with a patience of 15 epochs, and the checkpoint with the best validation loss is restored before evaluation. We use mean squared error for regression and binary cross-entropy with logits for classification.

\paragraph{Train--test distance.}
To quantify residual relatedness between the train and test sets, we compute the distance from each test sample to its nearest training sample using an exact brute-force search over the full training set. We use the same distance function as the corresponding splitting method: Tanimoto distance for molecular fingerprints and $1 -$ global sequence alignment identity~\citep{paper-global_alignment} for protein sequences. We report the empirical cumulative distribution function (ECDF) of these nearest-neighbor distances.

\subsection{Scaling}

\begin{figure}[h]
    \centering
    \begin{subfigure}{0.48\textwidth}
        \centering
        \includegraphics[width=\textwidth]{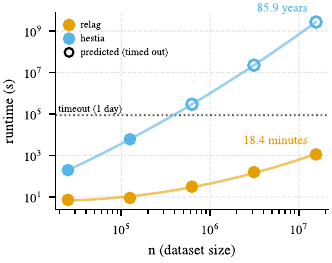}
        \caption{Runtime against dataset size on BELKA subsets, for \NAME and Hestia. Runs are capped at one day, and values above the dashed line are extrapolated from the fitted curve.}
        \label{fig:scaling_belka}
    \end{subfigure}
    \hfill
    \begin{subfigure}{0.48\textwidth}
        \centering
        \includegraphics[width=\textwidth]{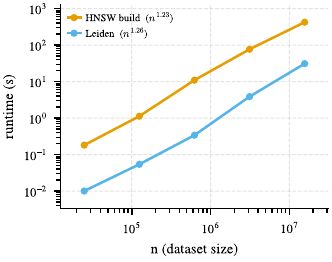}
        \caption{Per-stage breakdown of the \NAME pipeline on BELKA subsets. Building the HNSW dominates the total runtime at every size, but community detection grows faster ($n^{1.26}$ against $n^{1.23}$).}
        \label{fig:scaling_detailed}
    \end{subfigure}
    \caption{Scaling of \NAME on molecular modality. Both axes are logarithmic. Exponents in the legend are fitted on the log--log space.}
    \label{fig:scaling_appendix}
\end{figure}

We extend our scaling investigation to the molecule modality (Fig.~\ref{fig:scaling_belka}). Although \NAME scales substantially more favorably than Hestia, its empirical scaling appears slightly super-linear, above the expected $O(n \log n)$.

To investigate this discrepancy, we decompose the runtime of \NAME by algorithmic step and measure the contribution of each component as the dataset size increases (Fig.~\ref{fig:scaling_detailed}). This figure shows that the Leiden community detection seems to increase faster than the build process. This is expected because its computational complexity scales with the number of edges in the graph, which scales faster than $O(n\log n)$, suggesting that at very large scale, the runtime will be dominated by the community detection step.

\begin{figure}[t]
    \centering
    \begin{subfigure}{0.48\textwidth}
        \centering
        \includegraphics[width=\linewidth]{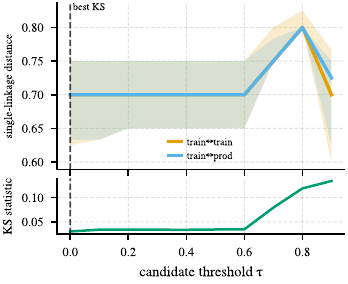}
        \caption{UniRef50, with ProteinGym, PEER and CASP15 as production data. The KS statistic is minimized at $\tau^\star = 0$ and remains nearly flat for $0 < \tau \lesssim 0.6$.}
        \label{fig:threshold_validation}
    \end{subfigure}
    \hfill
    \begin{subfigure}{0.48\textwidth}
        \centering
        \includegraphics[width=\linewidth]{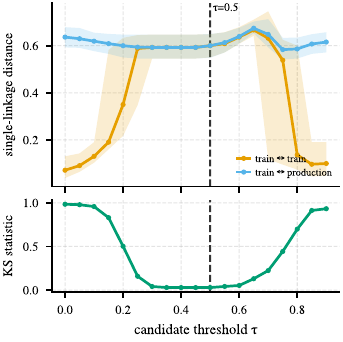}
        \caption{Synthetic peptides, with an independently generated production dataset. The KS statistic is minimized over a plateau, and the largest threshold on it, $\tau^\star = 0.5$, is selected.}
        \label{fig:synthetic_threshold}
    \end{subfigure}
    \caption{\textbf{Threshold selection on large-scale and synthetic data.} Upper panels: single-linkage distances from pure-training communities (train$\leftrightarrow$train) and from pure-production communities (train$\leftrightarrow$production) to training-containing communities, with shading denoting the 15th--85th percentile range. Lower panels: the corresponding Kolmogorov--Smirnov statistic. Dashed lines mark the selected threshold $\tau^\star$.}
    \label{fig:threshold_appendix}
\end{figure}

\subsection{Threshold Selection}
\subsubsection{Algorithm}
Algorithm~\ref{alg:threshold} details the threshold-selection procedure. Without prior knowledge of a suitable range, we sweep $\tau$ upward from $0$, where every sample is a singleton. The sweep stops when every candidate edge has been retained, as larger thresholds then leave the graph unchanged, or when no pure-training or pure-production communities remain, as the distance distributions are then undefined. Among the evaluated thresholds, we select the one minimizing the KS statistic, taking the largest such threshold in case of ties.
\begin{algorithm}[h]
\caption{Threshold selection}
\label{alg:threshold}
\begin{algorithmic}[1]
\Require Dataset to split $\mathcal{S}$, unlabeled production data $\mathcal{P}$, distance function $d$, increasing candidate thresholds $0 = \tau_1 < \tau_2 < \dots < \tau_K$
\Ensure Selected threshold $\tau^\star$ and minimal KS statistic $\mathrm{KS}^\star$
\State $E \gets$ base-layer edges of an HNSW graph built over $\mathcal{S} \cup \mathcal{P}$ with distance $d$
\For{$\tau \in (\tau_1, \dots, \tau_K)$}
    \State $E_\tau \gets \{(u, v) \in E : d(u, v) < \tau\}$ \Comment{Proximity graph at resolution $\tau$}
    \If{$E_\tau = E$}
        \State \textbf{break} \Comment{Complete graph}
    \EndIf
    \State $\gamma \gets p_0(\tau)$ \Comment{Null edge probability}
    \State $\Ctau \gets \textsc{Leiden-CPM}(\mathcal{S} \cup \mathcal{P}, E_\tau, \gamma)$
    \State $\Ctau_{\mathrm{train}} \gets \{A \in \Ctau : A \subseteq \mathcal{S}\}$ \Comment{Pure-training}
    \State $\Ctau_{\mathrm{prod}} \gets \{A \in \Ctau : A \subseteq \mathcal{P}\}$ \Comment{Pure-production}
    \State $\Ctau_{\mathrm{hybrid}} \gets \Ctau \setminus (\Ctau_{\mathrm{train}} \cup \Ctau_{\mathrm{prod}})$ \Comment{Hybrid}
    \If{$\Ctau_{\mathrm{train}} = \emptyset$ \textbf{or} $\Ctau_{\mathrm{prod}} = \emptyset$}
        \State \textbf{break} \Comment{Distributions undefined}
    \EndIf
    \State Compute $F^{(\tau)}_{\mathrm{train}\to\mathrm{train}}$ and $F^{(\tau)}_{\mathrm{prod}\to\mathrm{train}}$
    \State $k_\tau \gets \mathrm{KS}\big(F^{(\tau)}_{\mathrm{train}\to\mathrm{train}}, F^{(\tau)}_{\mathrm{prod}\to\mathrm{train}}\big)$
\EndFor
\State $\mathrm{KS}^\star \gets \min_\tau k_\tau$
\State $\tau^\star \gets \max\{\tau : k_\tau = \mathrm{KS}^\star\}$ \Comment{Largest threshold on ties}
\State \Return $\tau^\star, \mathrm{KS}^\star$
\end{algorithmic}
\end{algorithm}
\subsubsection{Uniref experiment}
We next apply the threshold-selection procedure at a scale representative of the training data used by state-of-the-art protein language models. We use the complete UniRef50 dataset, comprising approximately 38.7 million protein sequences. The production dataset is constructed as the union of three widely used protein benchmarks for evaluating protein language models: ProteinGym~\citep{paper-proteingym}, PEER~\citep{paper-peer}, and CASP15~\citep{paper-casp15}.

\begin{figure}[h]
    \centering
    \includegraphics[width=0.5\textwidth]{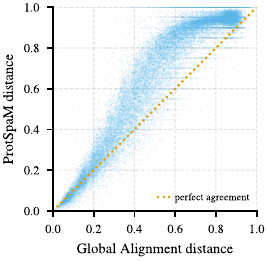}
    \caption{ProtSpaM distance against global alignment distance, on 39{,}380 pairs built from real UniRef50 seeds mutated at a sweep of rates (BLOSUM62 substitutions and indels) so the alignment axis is covered end to end. The two are monotonically related (Pearson $r = 0.954$, Spearman $r = 0.911$).}
    \label{fig:protspam_vs_alignment}
\end{figure}

At this scale, we replace global sequence alignment with the ProtSpaM kernel~\citep{paper-protspam} for distance computation. Global alignment has $O(L^2)$ complexity with respect to sequence length and is already the dominant computational cost on our smaller protein datasets, making its application to UniRef50 intractable. ProtSpaM instead estimates sequence distance from spaced-word matches, with approximately linear complexity in sequence length, making the threshold search feasible at this scale. This substitution does not substantially alter the interpretation of the distance: ProtSpaM and global alignment distances agree closely on the same sequence pairs (Fig.~\ref{fig:protspam_vs_alignment}), allowing thresholds expressed in ProtSpaM distance to be interpreted in terms of sequence identity.

The KS statistic reaches its minimum at $\tau=0$ and remains essentially flat after, for $\tau\in (0,0.6]$. It then increases for larger thresholds, reaching $0.080$ at $\tau=0.7$ and $0.135$ at $\tau=0.9$. The procedure therefore selects $\tau=0$.

This result is expected for two reasons. First, UniRef50 is clustered at 50\% sequence identity, equivalent to 0.6 in ProtSpaM distance (See Fig~\ref{fig:protspam_vs_alignment}), so relationships up to $\tau=0.6$ have already been removed by construction. Second, the production benchmarks have a similar distance to the training set as the training set is to itself (Table~\ref{tab:uniref_percentiles}). The production-to-training distribution is closer at the lower tail, with a first percentile of $0.00$ compared with $0.13$ for train-to-training, and remains closer in the upper tail at the 90th and 99th percentiles. Across the majority of the distribution, the two are very similar.

Thus, at small resolutions, the training and production samples are consistent with the same relatedness distribution, leading \NAME to correctly select $\tau=0$ and fall back to random splitting. This is consistent with the intended behavior of \NAME, as UniRef50 is pre-clustered to remove sequence redundancy, mitigating related-sequence leakage under random splitting. By contrast, in the BELKA challenge, the OOD production samples were too distant from the training distribution for any threshold to align the two distributions. These results suggest that PLM pretraining operates in a high-effective-sample-size regime: in our experiment, $N_{\mathrm{eff}}\approx N$ relative to widely used protein benchmarks, providing a potential explanation for why they generalize to multiple target tasks. BELKA represents the opposite regime, where aligning with the OOD production data requires a resolution at which $N_{\mathrm{eff}}$ collapses, providing a potential explanation for the poor generalization observed across models \citep{paper-del_iver_ood}.

\begin{table}[!htbp]
\centering
\small
\caption{Nearest-train-neighbor distance on UniRef50: percentiles of the distance from a training sequence to its closest other training sequence (train$\leftrightarrow$train), and from a production benchmark sequence to its closest training sequence (train$\leftrightarrow$prod).}
\label{tab:uniref_percentiles}
\begin{tabular}{rccc}
\toprule
Percentile & train$\leftrightarrow$train & train$\leftrightarrow$prod & prod $-$ train \\
\midrule
\textbf{1}  & \textbf{0.127} & \textbf{0.000} & $\mathbf{-0.127}$ \\
5           & 0.500 & 0.516 & $+0.016$ \\
10          & 0.600 & 0.600 & $0.000$ \\
25          & 0.650 & 0.650 & $0.000$ \\
50          & 0.700 & 0.700 & $0.000$ \\
75          & 0.750 & 0.750 & $0.000$ \\
\textbf{90} & \textbf{0.775} & \textbf{0.750} & $\mathbf{-0.025}$ \\
95          & 0.800 & 0.800 & $0.000$ \\
\textbf{99} & \textbf{0.825} & \textbf{0.817} & $\mathbf{-0.008}$ \\
\bottomrule
\end{tabular}
\end{table}

\subsection{Singletons: Noise or Signal?}
We investigate whether singleton samples contribute useful information or instead behave as outliers that reduce predictive performance. For each dataset, we train one MLP using only singleton training samples and a second MLP using an equally sized random sample of non-singleton training samples. Both models use the same embeddings, architecture, and the same fixed validation set obtained by a \NAME split, so the only difference is the subset of training samples used. We repeat the comparison over five splits with six repetitions per split and assess significance using a two-sided $t$-test with $p<0.05$.

To identify meaningful communities, thresholds are selected separately for each dataset using the heavy-tailed community-size heuristic of Section~\ref{sec:neff}. We use the modularity objective for all datasets, avoiding the need to recompute the CPM resolution parameter for each threshold and task.

The results reveal three regimes (Table~\ref{tab:singleton_power}). First, models trained on singletons outperform community-matched ones in \texttt{caco2\_wang}, \texttt{sr\_are}, and \texttt{lipophilicity}. This is consistent with singletons being genuine groups of size one: each would then carry independent information, whereas samples within a community share information with their relatives. Second, singletons are less informative in \texttt{ames}, \texttt{dbaasp}, and \texttt{cyp2c19\_veith} since models trained on singletons perform worse. This suggests that these singleton samples behave as outliers with a mapping that is less aligned with the prediction task. Finally, \texttt{pgp\_broccatelli} and \texttt{enzyme\_topt} show no statistically significant differences between singleton and community-matched samples, consistent with singletons representing a mixture of informative and uninformative samples.

These results indicate that whether singletons provide signal or noise is a property of the dataset rather than of the splitting method. We therefore exclude singletons from $N_{\mathrm{eff}}$ except when the statistical test indicates that they provide significantly greater per-sample signal than a community-matched sample, suggesting that they represent genuine groups rather than noise. Under this criterion, singletons contribute to $N_{\mathrm{eff}}$ for \texttt{caco2\_wang}, \texttt{sr\_are}, and \texttt{lipophilicity}, while they are excluded for the remaining five datasets (Fig.~\ref{fig:data-starvation_corr}).

\begin{table}[h]
\centering
\small
\caption{Singleton information content. Comparison of models trained on singleton samples and equally sized samples from non-singleton communities, evaluated on the same validation set. $\Delta$ is the performance difference (singleton $-$ community), with $\Delta>0$ indicating greater information content in singletons. Mean over 5 splits $\times$ 6 repeats. $p$-values are from two-sided $t$-tests. Rows are grouped by higher, lower, or indistinguishable singleton performance.}
\label{tab:singleton_power}
\begin{tabular}{llccccc}
\toprule
Dataset & Metric & $\tau$ & Singleton & Community & $\Delta$ & $p$ \\
\midrule
\texttt{caco2\_wang}      & PCC   & 0.6 & $0.473 \pm 0.122$ & $0.417 \pm 0.103$ & $+0.056$ & $1.1\times10^{-2}$ \\
\texttt{lipophilicity}    & PCC   & 0.4 & $0.540 \pm 0.037$ & $0.500 \pm 0.054$ & $+0.040$ & $2.6\times10^{-5}$ \\
\texttt{sr\_are}          & AUROC & 0.4 & $0.710 \pm 0.030$ & $0.682 \pm 0.030$ & $+0.029$ & $1.5\times10^{-6}$ \\
\midrule
\texttt{cyp2c19\_veith}   & AUROC & 0.5 & $0.779 \pm 0.045$ & $0.796 \pm 0.020$ & $-0.017$ & $3.8\times10^{-2}$ \\
\texttt{dbaasp}           & PCC   & 0.3 & $0.483 \pm 0.061$ & $0.520 \pm 0.041$ & $-0.037$ & $1.3\times10^{-5}$ \\
\texttt{ames}             & AUROC & 0.7 & $0.530 \pm 0.075$ & $0.640 \pm 0.064$ & $-0.109$ & $8.8\times10^{-5}$ \\
\midrule
\texttt{pgp\_broccatelli} & AUROC & 0.6 & $0.817 \pm 0.095$ & $0.830 \pm 0.055$ & $-0.013$ & $0.382$ \\
\texttt{enzyme\_topt}     & PCC   & 0.8 & $0.573 \pm 0.103$ & $0.602 \pm 0.108$ & $-0.029$ & $0.175$ \\
\bottomrule
\end{tabular}
\end{table}

\FloatBarrier 
\subsection{Additional figures}
\begin{figure}[h]
    \centering
    \includegraphics[width=0.45\linewidth]{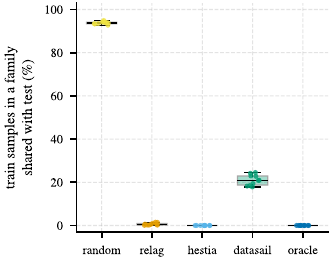}
    \caption{\textbf{Family leakage on the synthetic dataset.} Fraction of training samples whose family also appears in the test subset. Random: 93.8\%, \NAME: 0.62\%, DataSAIL: 21.01\%, Hestia: 0.00\%, Oracle: 0.00\%}
    \label{fig:placeholder}
\end{figure}
\begin{figure}[h]
    \centering
    \begin{subfigure}{0.45\textwidth}
        \centering
        \includegraphics[width=\textwidth]{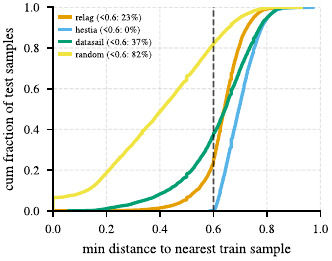}
        \caption{Small-molecule datasets}
        \label{fig:leakage-molecule}
    \end{subfigure}
    \hfill
    \begin{subfigure}{0.45\textwidth}
        \centering
        \includegraphics[width=\textwidth]{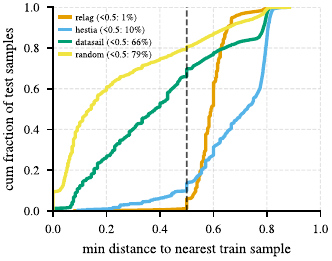}
        \caption{Protein-sequence datasets}
        \label{fig:leakage-protein}
    \end{subfigure}
    
    \caption{
        \textbf{Train--test distance across data-splitting methods.}
        ECDFs of the minimum distance from each test sample to its nearest training sample, pooled within modality: \textbf{(a)} six small-molecule datasets and \textbf{(b)} two protein-sequence datasets. Distance is $1-$ similarity, using Tanimoto similarity for molecules and global sequence identity~\citep{paper-global_alignment} for proteins. Dashed lines indicate the splitting thresholds ($0.60$ and $0.50$, respectively); the legend reports the fraction of test samples closer to any train samples than the corresponding threshold.
    }
    \label{fig:leakage-ecdf}
\end{figure}

\begin{figure}[h!]
    \centering
    \begin{subfigure}{0.48\textwidth}
        \centering
        \includegraphics[width=\textwidth]{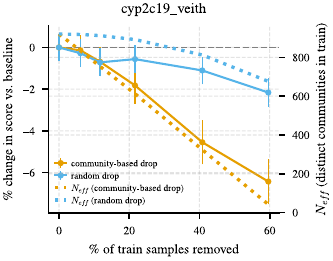}
    \end{subfigure}
    \hfill
    \begin{subfigure}{0.48\textwidth}
        \centering
        \includegraphics[width=\textwidth]{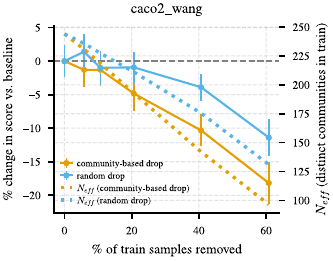}
    \end{subfigure}
    \begin{subfigure}{0.48\textwidth}
        \centering
        \includegraphics[width=\textwidth]{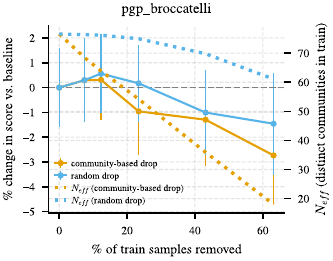}
    \end{subfigure}
    \hfill
    \begin{subfigure}{0.48\textwidth}
        \centering
        \includegraphics[width=\textwidth]{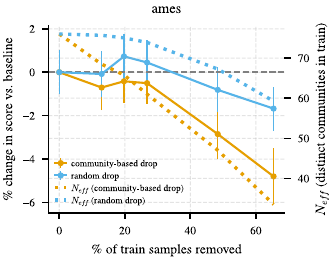}
    \end{subfigure}
    \begin{subfigure}{0.48\textwidth}
        \centering
        \includegraphics[width=\textwidth]{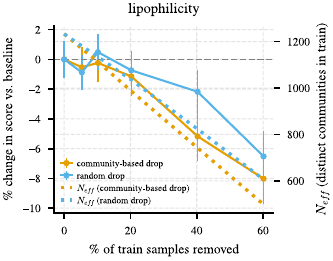}
    \end{subfigure}
    \hfill
    \begin{subfigure}{0.48\textwidth}
        \centering
        \includegraphics[width=\textwidth]{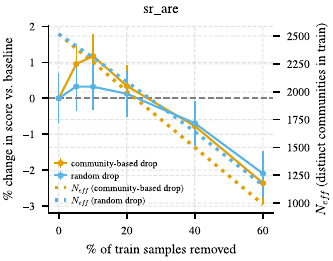}
    \end{subfigure}
    \begin{subfigure}{0.48\textwidth}
        \centering
        \includegraphics[width=\textwidth]{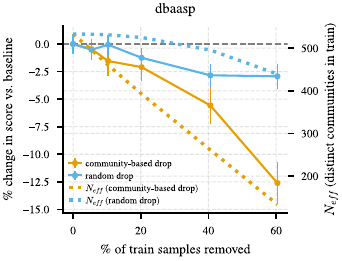}
    \end{subfigure}
    \hfill
    \begin{subfigure}{0.48\textwidth}
        \centering
        \includegraphics[width=\textwidth]{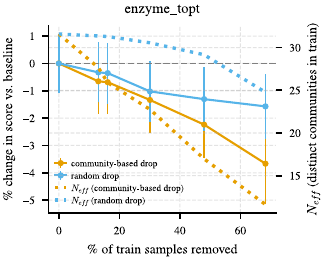}
    \end{subfigure}
    \caption{Effect on test performance of dropping random samples or full communities on different datasets. $N_{\mathrm{eff}}$ corresponds to the number of communities remaining in the dataset.}
    \label{fig:data_starv_supp}
\end{figure}
\end{document}